# A Method of Fluorescent Fibers Detection on Identity Documents under Ultraviolet Light

Kunina I.A.[1,3,4], Aliev M.A.[2,3], Arlazarov N.V.[3], Polevoy D.V.[2,4,5]
[1]Institute for information transmission problems (Kharkevich Institute) RAS, Moscow, Russia
[2]Federal Research Center "Computer Science and Control" RAS, Moscow, Russia
[3]Smart Engines Service LLC, Moscow, Russia
[4]Moscow Institute of Physics and Technology, Dolgoprodny, Russia
[5]National University of Science and Technology "MISIS", Moscow, Russia

## ABSTRACT

In this work we consider the problem of the fluorescent security fibers detection on the images of identity documents captured under ultraviolet light. As an example we use images of the second and third pages of the Russian passport and show features that render known methods and approaches based on image binarization non applicable. We propose a solution based on ridge detection in the gray-scale image of the document with preliminary normalized background. The algorithm was tested on a private dataset consisting of both authentic and model passports. Abandonment of binarization allowed to provide reliable and stable functioning of the proposed detector on a target dataset.

## 1. INTRODUCTION

At present the OCR systems are increasingly used for ID documents [1-3]. Although usually the main purpose of such systems is automation of personal data input, automatic verification of the document authenticity also becomes more and more popular.

To determine whether the document is authentic the system can check the contents of the recognized document image for compliance with certain rules: presence of holographic elements in specific places [4,5], stamps [6,7] of certain types, usage of standardized fonts [8,9], etc. Besides that, some checks can be run on the entire image: for example, system can refuse to process the recognized personal data if it is revealed that the image was edited [10,11] or has very poor quality [12].

Modern compact semiconductor light emitters and small-scale digital cameras allow to create small low-cost full page document scanners without moving parts. In such scanners light emission is carried out by one or several LEDs, and the camera is placed in the lower part opposite to the working surface made of transparent glass [13,14]. Such scanners are successfully used in ATMs and self service booths, or access control systems at checkpoints.

Using different types of LEDs allows to create document images in different spectral bands (for example, visible light, IR and UV), which in turn increases the number document authenticity features that can be verified.

One of such features is the presence of fibers fluorescing in UV light which are often used in manufacturing of different types of documents. Among them are Russian passports (see Fig. 1), documents of European Union (EU) citizens according to PRADO (Public Register of Authentic travel and identity Documents Online) [15] and others.

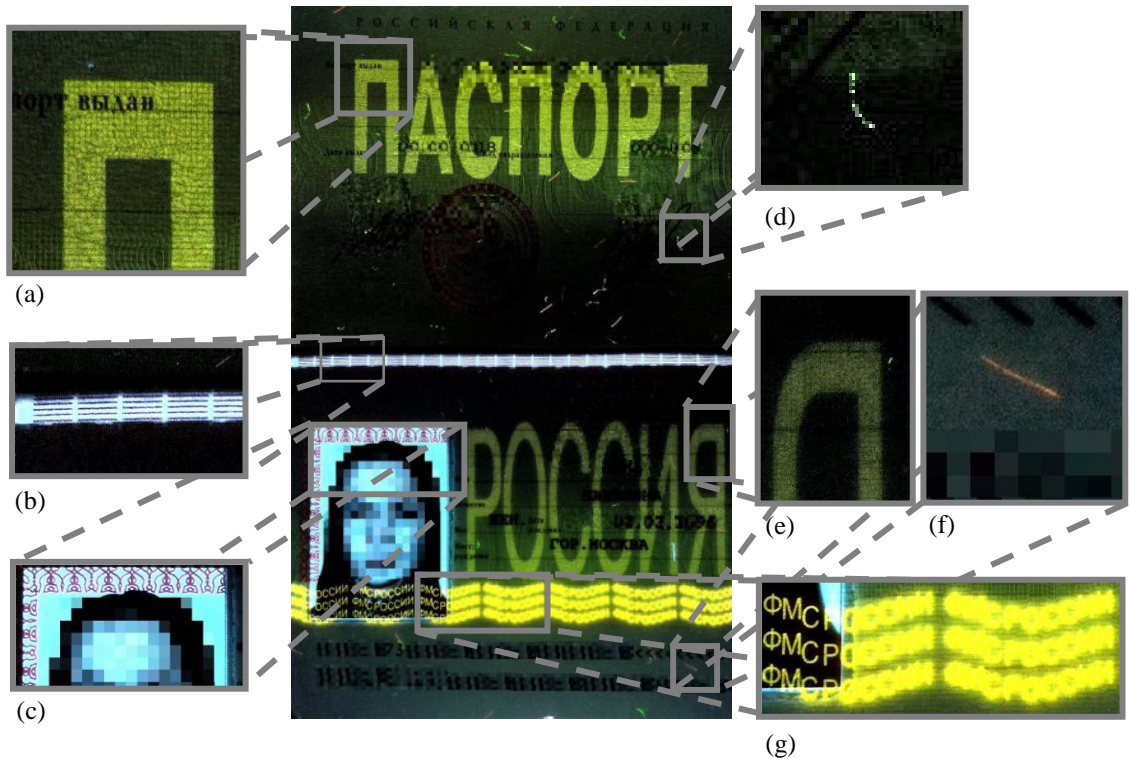

Figure 1. Examples of the second and third pages of the Russian passport in UV light. The following security elements are fluorescent in UV light: (a) - the word russian «ПАСПОРТ» on the second page, (b) - the strip that divides the second and third pages, (d) and (f) - security fibers embedded in the paper substrate, (e) - the word russian «ФМС РОССИИ» on the third page (also possible are allowed variations: russian «МВД РОССИИ» or russian «РОССИЯ»). The owner photo (c) may have higher level of luminosity compared to the document background.

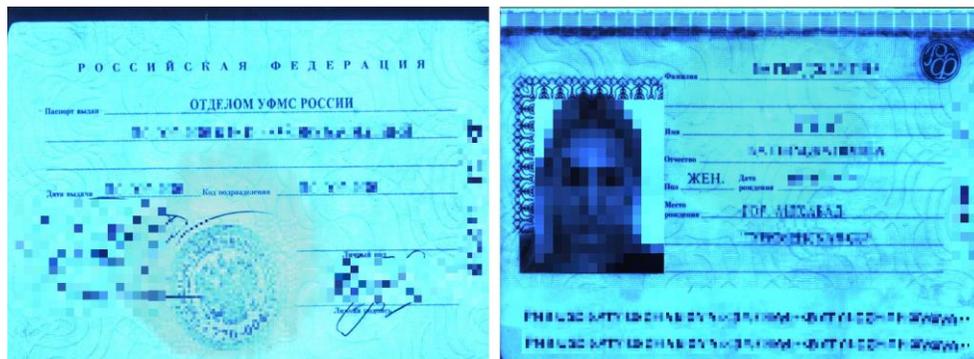

Figure 2. Examples of the second and third pages of a model Russian passport in UV light: absence of security elements that should be fluorescent in UV light, high background luminosity of the whole document.

In this paper we consider the problem of detecting the fluorescent security fibers in the images of identity documents captured under ultraviolet light in order to evaluate document authenticity. By security fibers here we mean thin short multicolored fibers, randomly located in the paper substrate (see Fig. 1d,f). We assume that in the input image only one document page is present, projective distortions are absent, and the document page covers the whole image (this is achieved by localization of the document page zone using recognition methods described in [2]).

The proposed algorithm was tested on images of the second and third pages of Russian passports, both authentic (see Fig. 1) and model (see Fig. 2).

## 1.1 Features of Russian passports under UV light

Using special materials based on paper and plastic [16-18] for printing identity documents is a worldwide practice. Pages made from such materials have low luminosity in UV light, while standard commercial types of paper glow under such light with bright blue color as shown in Fig. 2. Additional security level is achieved by various fluorescent inclusions in paper mix: fibers, filaments, confetti. Active development of different new variants of such inclusions has been carried out for many years (for example [19,20]) and still continues today [21]. Nevertheless, fluorescent fibers continue to be one of the most utilized types of security inclusions.

In Russia, in compliance with the technical standards for printing protected documents [18], paper for such production must incorporate fibers of at least two colors (see Fig. 1d,f). Besides different colors such fibers often have different levels of brightness along the fiber due to varying levels of immersion in the paper mix during its production (see Fig. 1d). To make fluorescent inclusions even more contrasting the paper mix must absorb most of the light. However, imperfections of scanner devices may lead to some luminescence of the background paper, or excess exposure can occur by mistakes during scanner setup process (incorrect calibration) or document scanning (document not being pressed hard enough to the glass). In Fig. 3 are presented some examples of security fibers on the document images that were used for testing of the proposed algorithm. There can be seen varying levels of background and security fibers luminosity in different documents, as well as that not only luminosity differs, but also the background color. Also, in UV light there may be fluorescence from the security elements other than security fibers (see Fig. 1a,b,e,g), and the owner photograph may have higher level of luminosity (see Fig. 1c).

All this should be taken under consideration during development of the security fibers detection algorithm on identity documents images.

## 1.2 Overview of fibers detection methods

For authenticity check of banknotes it is suggested in [22] to detect and classify fluorescent fibers in UV light. Detection is realized in the following way: first, the image is converted to gray-scale (taking in consideration a dominant color of fluorescent elements). Next, Otsu global binarization method is used and then geometric analysis of connected components. Because of the large number of false positives the image fragments in detection regions of interest are classified using a neural network.

In work [23] a system for automation of carcinogen (asbestos) fibers count using micro-photographs analysis of special filters is described. The proposed approach consists of the following steps: background subtraction to compensate for non-uniform lightness of a sample with the use of unpolluted filter image and taking into account possible presence of fluorescent inclusions; binarization; calculation of connected components; connected components analysis specific for task of particles count (contouring, axial line estimation, proportion estimation, special branches and intersections processing; count of objects with given geometric properties).

In [24] a similar counting scheme in micro-photograph is used for glass fibers, but additionally the Contrast Limited Adaptive Histogram Equalization is used to increase the image contrast and then noise suppression with adaptive radial convolution filter is applied.

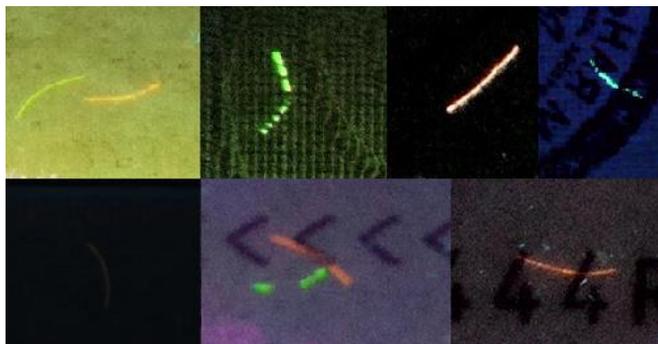

Figure 3. Examples of fluorescent security fibers in different images of Russian passports from the test dataset

Our task in this work is further complicated by a number of factors illustrated in the Fig. 1: large number of security features, high background luminosity of owner photograph, text elements of the document. Small number of light sources in scanners without moving parts adds one more factor to the difficulty - highly varied illumination of the document page: in the Fig. 1 it can be observed that the background around the word russian «ПАСПОРТ» is lighter than the part under the word, and in the Fig. 3 the fibers and background fragments have different color and lightness. All these aspects make the binarization methods non applicable.

In this work we propose to detect fibers as sets of points, each of which forms a ridge on a 1D profile of brightness in brightness gradient direction. For suppression of false positives the image is processed with morphological filtration and background normalization.

## 2. LOCALIZATION OF FIBERS IN THE IMAGE

Because security fibers have different colors the image acquired from scanner is first converted to gray-scale by taking an average value of the channels in each pixel. From here on by image we will mean such a gray-scale image.

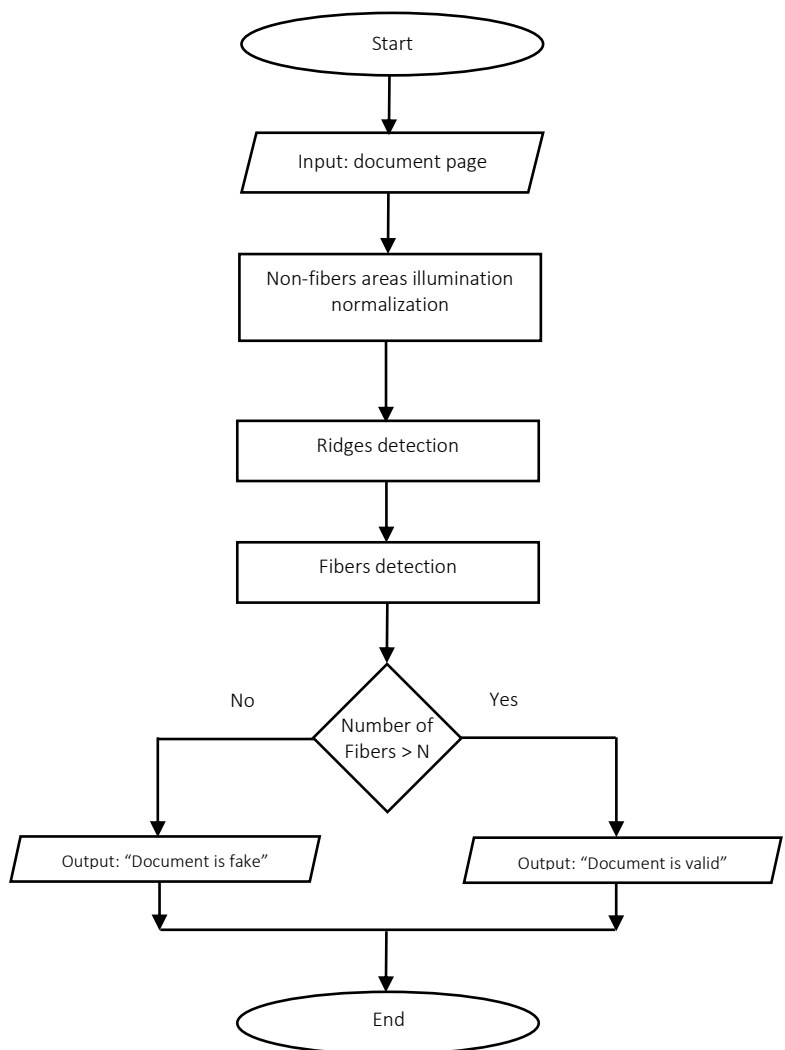

Figure 4. Control flowchart of the security fibers localization in the document image and decision-making about document authenticity

### 2.1 Localization of points that form ridge structures

Let us introduce the following model of a point that belongs to a fiber image: it is a point that forms local maximum (ridge) in the source document image brightness profile in brightness gradient direction. Most popular criterion for checking if selected point is described by the introduced model is the criterion [25], based on whether the Hessian matrix first eigenvalue is negative. However, using this criterion leads to response not only on the ridges but also on objects boundaries that are wider that the fibers.

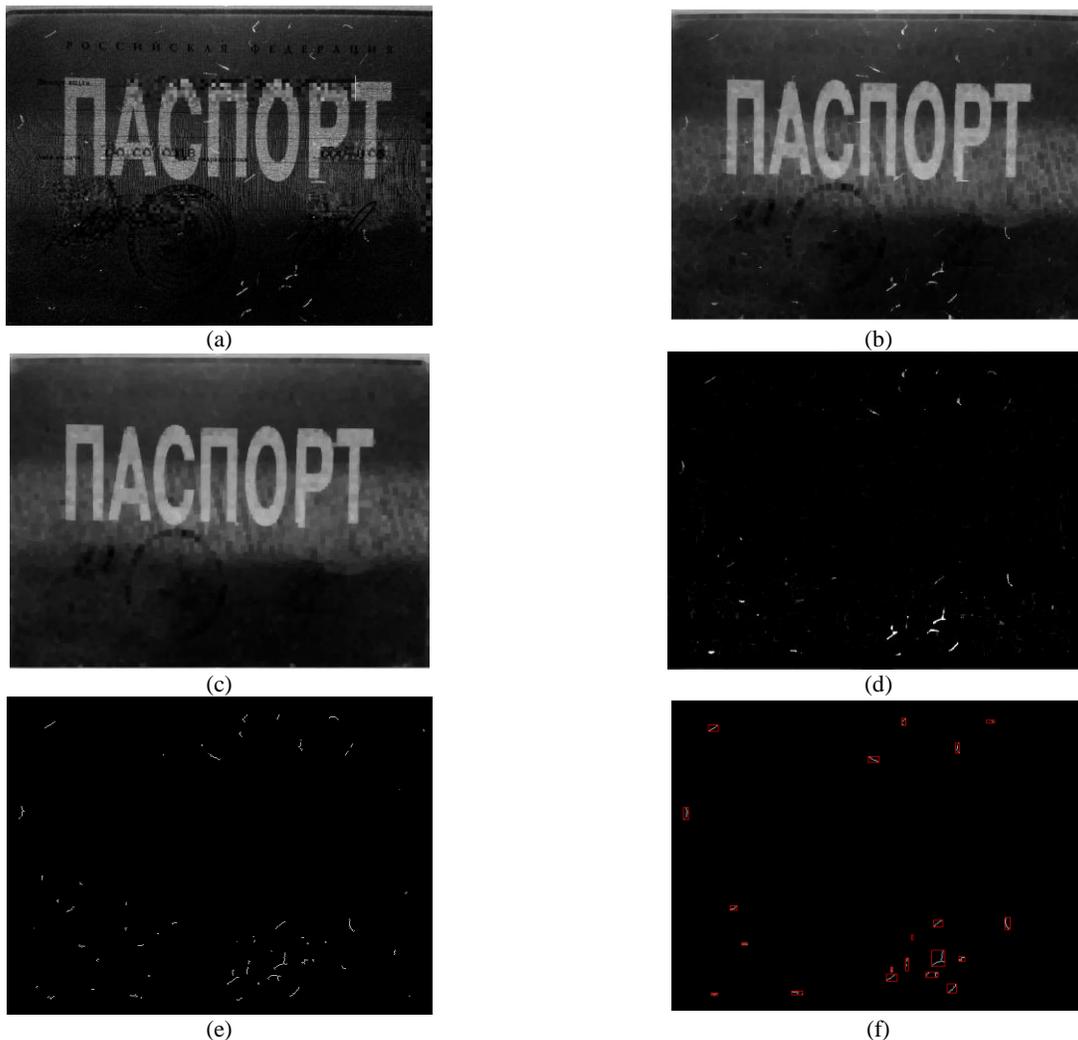

Figure 5. Example of security fibers detection in the input image (a). Main stages of algorithm are shown: (b) – input image text suppression; (c) – application of opening morphological operation (background acquiring); (d) – document image with normalized background as the result of division pixel-by-pixel for the image b and image c dilated by square structuring element; (e) – the result of ridges detection as described in 2.1, (f) – localization and analysis of connected components. Red rectangles correspond to found security fibers.

That is why in this work to check if a point belongs to a ridge it was decided to compare brightness values in the source image near the point in the eigenvector direction, which corresponds to second eigenvalue of the Hessian, which corresponds to the gradient direction. To increase the accuracy of determining the ridge direction we used Scharr differential operator [26], which has rotational symmetry, during the Hessian calculation. Points that have passed this check were further passed through the stages of "Non-maximum Suppression" and "Hysteresis Thresholding" similar to those used in Canny edge detector [27]. Suggested approach of ridges detection was successfully used in the vanishing points detection task as the first stage of setting the eigen coordinate system in a projectively distorted document image [28].

To increase detection accuracy, text information was first suppressed in the input image and then the acquired image was passed to a background normalization stage. This was done because the gaps between the letters in a small area also satisfy the ridge model introduced above and fibers brightness may vary a lot both in a single document and in the whole dataset which makes the choice of the threshold values difficult.

## 2.2 Text suppression and background normalization

Initially all text information is suppressed in the input image. For this purpose the morphological operation of closing using a square structure element is applied to the input image (see Fig. 5a). As a result the document image $I_0$ with suppressed text information is acquired (see Fig. 5b).

Next, the morphological operation of opening using a square structure element is applied to the image $I_0$ for acquiring an image of document background $I_{bg}$ (see Fig. 5c). Now, to normalize the background of the image $I_0$ we, first, apply dilatation operation using square structure element to the image $I_{bg}$ and acquire the image $I_{dil}$, and then perform pixel-by-pixel division for the images $I_0$ and $I_{dil}$ (see Fig. 5d). In addition, in the case of in the third passport page image, the pixel values in the approximate region of the owner photograph in the obtained image is set to one.

After that, ridge points are localized (see Fig. 5e) on the result image as described above.

## 2.3 Fibers localization and decision on the document authenticity

The result of ridge points localization is a binary map where nonzero pixel is classified as a fiber pixel (see Fig. 5e). Next, the acquired pixels are united in connected components, where the allowed distance between two pixels is set by taking into account a possible break of fiber image continuity in the document page.

The resulting connected components are filtered by length, where length is defined as a quantity of pixels in the component. In the Fig. 5f found and checked connected components are selected with red rectangles.

The document is accepted as authentic if the quantity of connected components is higher than minimum allowed number of fibers in the image, otherwise the document is considered as fake.

The general form of the proposed algorithm is shown in Fig. 4. Here the algorithm input is a gray-scale document page image acquired in UV light and the minimum allowed number of fibers in the image. The background of the input image is normalized. Then, in the acquired image the points are localized that form ridges on image brightness profile in brightness gradient direction. The found points are united in groups. Groups that passed the check are considered as found fibers. The document is accepted as authentic if number of found fibers is more than the given minimal value.

## 3. TESTING OF THE PROPOSED ALGORITHM

The algorithm was tested on the dataset that contains 95 images of the second page and 2582 images of the third page of Russian passports, out of which 34 and 42 images respectively are model passports scans. Images were acquired using PS4-02 [14] scanner developed by Intek group. This dataset has a high variability of security fibers luminescent properties and background luminosity (see Fig. 3).

Detection result precision was 99.4%, recall was 100%.

What is important is that the algorithm successfully copes with the model passports considering the infrequency and probable losses of missing a fake. From the other side authentic document recognized as a fake might be passed for the final decision to operator.

The drawbacks of the algorithm can include missing the fibers in regions near other fluorescent security elements (which is caused by features of chosen preprocessing). Also the algorithm might miss the fibers because of their weak fluorescent properties or because of fibers weak visibility in case of poor lighting.

## 4. FURTHER RESEARCH

Specialized document scanners based on small format digital cameras allow in a short time to acquire several document images with different capture conditions. In that case an improvement of detection result can be achieved, for example, by creating a series of images made with different exposure times and combining the detection results for all captured images using the mathematical model of symbol recognition in a video stream [29]. Another option for increasing the detection quality might be the combination of images captured with different light conditions [30].

Authors also plan to widen the test dataset by adding documents of another types and countries, for example European Union citizen documents in accordance with PRADO [15].

## 5. CONCLUSION

In this work we proposed a new method for luminescent security fibers detection in identity document images under UV light. Method novelty is abandonment of binarization, which is the base of existing approaches, and applying instead a ridge structures analysis and specialized input image preprocessing for noise suppression.

Algorithm testing was carried out in images of the second and third pages of Russian passports from a private dataset containing both authentic and model documents. Detection result precision is 99.4%, recall is 100%.

Regardless of the specificity of the test dataset the proposed algorithm is general and can be effectively applied for security fibers detection on other types of documents because luminescent security fibers have common properties for different types of documents.

## ACKNOWLEDGMENTS

This work is partially financially supported by Russian Foundation for Basic Research (projects 17-29-03370 and 17-29-03263).